\documentclass{article}
\usepackage{times}
\usepackage{epsfig}
\usepackage{graphicx}
\usepackage{amsmath}
\usepackage{amssymb}
\usepackage{amsfonts} 
\usepackage[normalem]{ulem}
\usepackage[table,xcdraw,dvipsnames]{xcolor}
\usepackage{colortbl}

\usepackage{color,soul}  
\definecolor{ao(english)}{rgb}{0.0, 0.5, 0.0}
\def\etal{\emph{et al.\ }}

\usepackage[final,nonatbib]{neurips_2022}

\usepackage[utf8]{inputenc} 
\usepackage[T1]{fontenc}    
\usepackage{hyperref}       
\usepackage{url}            
\usepackage{booktabs}       
\usepackage{amsfonts}       
\usepackage{nicefrac}       
\usepackage{microtype}      
\usepackage{xcolor}         

\title{End-to-End Multimodal Representation Learning for Video Dialog}

\author{%
    Huda Alamri \\
    Georgia Institute of Technology \\
    \texttt{halamri3@gatech.edu} 
    \And
    Anthony Bilic \\
    Georgia Institute of Technology \\
    \texttt{abilic3@gatech.edu}
    \And 
    Michael Hu \\
    Georgia Institute of Technology \\
    \texttt{mhu93@gatech.edu}
    \And
    Apoorva Beedu \\
    Georgia Institute of Technology \\
    \texttt{abeedu3@gatech.edu}
    \And 
    Irfan Essa  \\
    Georgia Institute of Technology \\
    \texttt{irfan@gatech.edu}
} 

\begin{document}

\maketitle

\begin{abstract}
Video-based dialog task is a challenging multimodal learning task that has received increasing attention over the past few years with state-of-the-art obtaining new performance records. This progress is largely powered by the adaptation of the more powerful transformer-based language encoders. Despite this progress, existing approaches do not effectively utilize visual features to help solve tasks. Recent studies show that state-of-the-art models are biased towards textual information rather than visual cues. In order to better leverage the available visual information, this study proposes a new framework that combines 3D-CNN network and transformer-based networks into a single visual encoder to extract more robust semantic representations from videos. The visual encoder is jointly trained end-to-end with other input modalities such as text and audio. Experiments on the AVSD task show significant improvement over baselines in both generative and retrieval tasks.

\end{abstract}

\section{Introduction}
\label{sec:intro}

\begin{figure}[!ht]
    \centering
     \includegraphics[width=0.8\columnwidth]{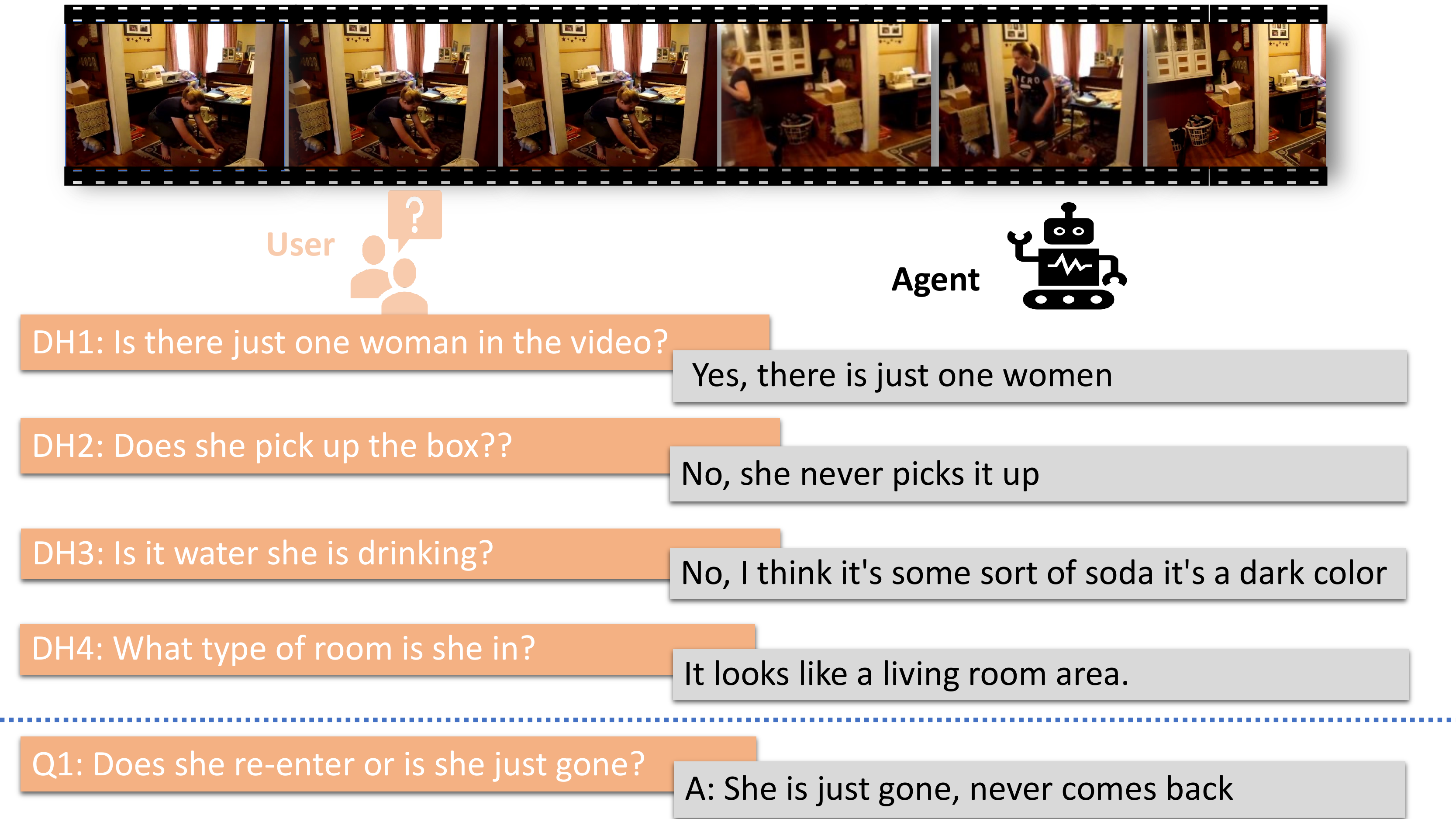}
    \caption{In  video dialog task, the model is presented with a short video, a dialog about the video, and a follow-up question. The goal is to correctly answer the question conditioned on the audio-visual cues and the dialog history (DH).}
    \label{fig:AVSD_example}
\end{figure}

The goal of the video-based dialog task is to answer questions about a dynamic scene presented in the video. More precisely, given a short video clip and multiple rounds of questions and answers about the video, the model should provide an accurate response to a follow-up question. An example of this is shown in ~\ref{fig:AVSD_example}, where a model is presented with a short video and a conversation about it. When the model is asked a follow-up question: \textit{“Did she re-enter the room?”}, to provide an accurate answer, the model has to acknowledge that the person “she” refers to the “woman” mentioned in the previous utterances. The model also has to identify the action “re-entering the room” from the actions in the video. 
This video-based dialog task represents a challenging multi-modal learning problem that serves as a test bed for video and language representation learning. Advances in this research field influences a wide range of applications, including providing road assistance for autonomous vehicles ~\cite{kim2018textual}, helping visually impaired individuals to understand their surroundings, and navigating through a very long video etc.

Success in this multi-modal learning task hinges on tackling four main challenges: \emph{(i)} extracting strong visual representations; \emph{(ii)} extracting strong textual representations; \emph{(iii)} effectively combining both features with other modalities (audio, when available); and finally, \emph{(iv)} generating an accurate response in natural language. While the task has received considerable interest from the community, current work largely focuses on obtaining strong textual and visual representations independently and combining the features ~\cite{bao2019plato,videoGPT2,schwartz2019simple,lee2020dstc8,le2019multimodal}, while the knowledge and cues from the video-text association have not been extensively explored. This was investigated by Liu~\etal\cite{liu2022revisiting}, who demonstrated that most models are biased towards the textual information, while visual features not contributing substantially towards performance. This study argues that using the visual features extracted from frozen 3D-CNN networks learned from action recognition data, without the added knowledge about the corresponding text association, i.e. the questions, result in reduced performance compared to joint training with both modalities.

Our work addresses this limited utilization of visual information in the video-based dialog task by making the models more visually aware. 
First, a $3$D-CNN network extracts local temporal features from the input video, which is then passed to a transformer based visual encoder network that generates contextual representation through self-attention mechanism. These visual features are then effectively combined with text and audio features to generate a best response for the input video and question. These multiple modules form one unified framework that is trained end-to-end which enables the model to generate stronger latent representations. Experiments on the video-based dialog task AVSD show that our model learns a stronger joint visual-textual features which contribute significantly to its performance.
Through several baselines, we show how recent methods pre-extract visual features and improve the vision-based language tasks due to the strong performance of the language models (e.g.,BERT and GPT2). On the contrary, our framework is designed to use standard architectures to emphasize that joint learning of visual and textual information is vital for the video-dialog task.

The contributions of our work are as follows:

\vspace{-5pt}
\begin{itemize}
    \item We propose a new framework for video-based language understanding and generation tasks. 
    This multi-modal framework effectively learns contextual representation using strong visual features from video, and through self-attention.   
    \vspace{-5pt}
    \item Our framework is flexible and can use any number of modalities and different encoders for these inputs. We show ablations on using Audio in addition to Text and Video modalities in~\ref{sec:experiments}.
    \vspace{-5pt}
    \item We also show the effectiveness of joint training on the retrieval task with a simpler framework. We provide extensive experiments and detailed analysis of both generative and retrieval tasks in the AVSD dataset are provided.

\end{itemize}

\section{Related Work}
\label{sec:related_work}

\vspace{-0.1in}
Video and language understanding has been extensively investigated due to the wide range of potential applications in human-computer interactions. 
Tasks such as video captioning~\cite{wang2018reconstruction,zhou2018end,chen2018tvt}, video question-answering~\cite{lei2018tvqa,yang2003videoqa,park2021bridge,lei2019tvqa+}, and video dialog ~\cite{alamri2019audio,jin2019video,li2021bridging} study the complex interplay between the vision and natural language modalities. In the case of video question-answering, effective performance depends on extracting strong visual representation for the input video and efficiently fusing it with the associated text. 
For video dialog, Alamri~\etal~\cite{alamri2019audio} introduced the Audio-Visual Scene Aware Task (AVSD) as a multi-modal learning problem, the objective of which is to answer a question based on a short video, with an associated audio and a dialog history. The task supports a discriminative setting, where the model ranks a list of candidate answers ~\cite{alamri2019audio,visdialbert}, or a generative setting, where a decoder is trained to auto-regressively generate an answer~\cite{videoGPT2,hori2019end}.

Self-attention models, known as transformers~\cite{attention}, have been very successful at generating deep contextual linguistic representations. They are generally pre-trained with self-supervised learning on very large unlabelled text corpora, and subsequently fine-tuned on downstream tasks. They deliver state-of-the-art results for several natural language understanding and generation tasks~\cite{attention,GPT2,ELMO,bert18}. In our work we utilize a pre-trained BERT\cite{bert18} model to encode the input question and the dialog history. 

Inspired by this success, a large body of work has adapted self-attention models to multi-modal learning, including image question answering~\cite{vilbert,visdial,VisualBert,unicoder,lxmert}, image dialog ~\cite{visdial}, video question answering~\cite{videobert,sun2019contrastive,videobert}, and video dialog~\cite{videoGPT2,bao2019plato,le2019multimodal}.
In general, these approaches can be categorized into single-stream and two-stream networks. 

In the two-stream approach, each modality is independently encoded by a transformer-based network, and information is fused through concatenation or cross-attention ~\cite{lxmert,vilbert}. In the one-stream approach, Li~\etal\cite{VisualBert}, 
Su~\etal\cite{vl}, and  Li~\etal\cite{visdialbert} utilize a unified transformer network where video and text tokens are combined as one sequence. 

In the two-stream approach, the visual features and the text features extracted using modal specific encoders then fused jointly via transformer-based encoder Luo~\etal\cite{univl}. This study builds on the proposed model in~\cite{univl} and extends it in two ways: first, a 3D-CNN network is added to the backbone visual encoder. Second, an audio transformer-based encoder is added to learn a representation from the audio signal, which is combined with the other modalities via a cross-encoder, and the different encoders and the decoder are jointly trained in an end-to-end fashion. The experiments demonstrate the benefits of this approach.

Le. H.~\etal proposed a multimodal transformer network with query-attention~\cite{le2019multimodal}. Zekang et al.~\etal\cite{GPT2} utilized a pretrained GPT$2$ model and extended it to learn joint audio\-visual and text featuring by training the model on multi\-task learning objectives \cite{videoGPT2}. Cherian A.~\etal extend the audio-visual transformer by adding student-teacher learning~\cite{shah2021audio}. While all these approaches for video dialog tasks have achieved promising improvements, the utilization of visual features remains limited. All the approaches rely on pre-extracted visual features from 3D-CNN networks with no further fine-tuning or training. This has resulted in models that do not fully capture the multimodal nature of the task~\cite{liu2022revisiting}. In contrast, this model designed in this study also updates the visual extractor (a 3D-CNN) in an end-to-end fashion, which leads to the improved learning of visual features tailored to the video question answering task.

\section{Method}
\label{models}
This section introduces the framework for the video-based dialog task. It presents the different modal-specific encoders, pre-processing of the input modalities, training objectives, and the evaluation process.

\vspace{-1em}
\subsection{Task Formulation} 
Given an input video ${V}{=} (V_1,\dots,V_i,\dots,V_n)$, where $V_{i}$ is the $i^{th}$ frame sampled from the video, a dialog history ${{DH}_t {=}(C,(Q_1, \textsl{Ans}_1),\cdots},{(Q_{t-1},\textsl{Ans}_{t-1}))}$, where $C$ is the video caption and $(Q_{t-1},\textsl{Ans}_{t-1})$ corresponds to a question-answer pair at round $t-1$, and audio $A$ (see Figure~\ref{fig:AVSD_example}), the task is formulated such that, given a follow-up question $Q_t$, the model must generate a response $R_t$ considering input features: $V$, $DH_{1:(t-1)}$, $A$, and $Q_{t}$:

\vspace{-0.5em}
\begin{equation} \label{eq1}
    P(R_t|V,A,DH_{1:(t-1)},Q_t;\theta) = \prod_{j=0}^{t-1} P(R_j|V,A,DH_{1:j-1},Q_t;\theta)
\end{equation}
and train to minimize the cross entropy loss: 
\vspace{-0.5em}
\begin{equation} 
    \label{eq1}
    \mathcal{L}(\theta) = -\log P(R_t|V,A,DH_{1:(t-1)},Q_t;\theta)
\end{equation}
where $\theta$ comprises of the trainable network parameters.

\begin{figure*}[t]
    \centering
    \includegraphics[width=0.9\textwidth]{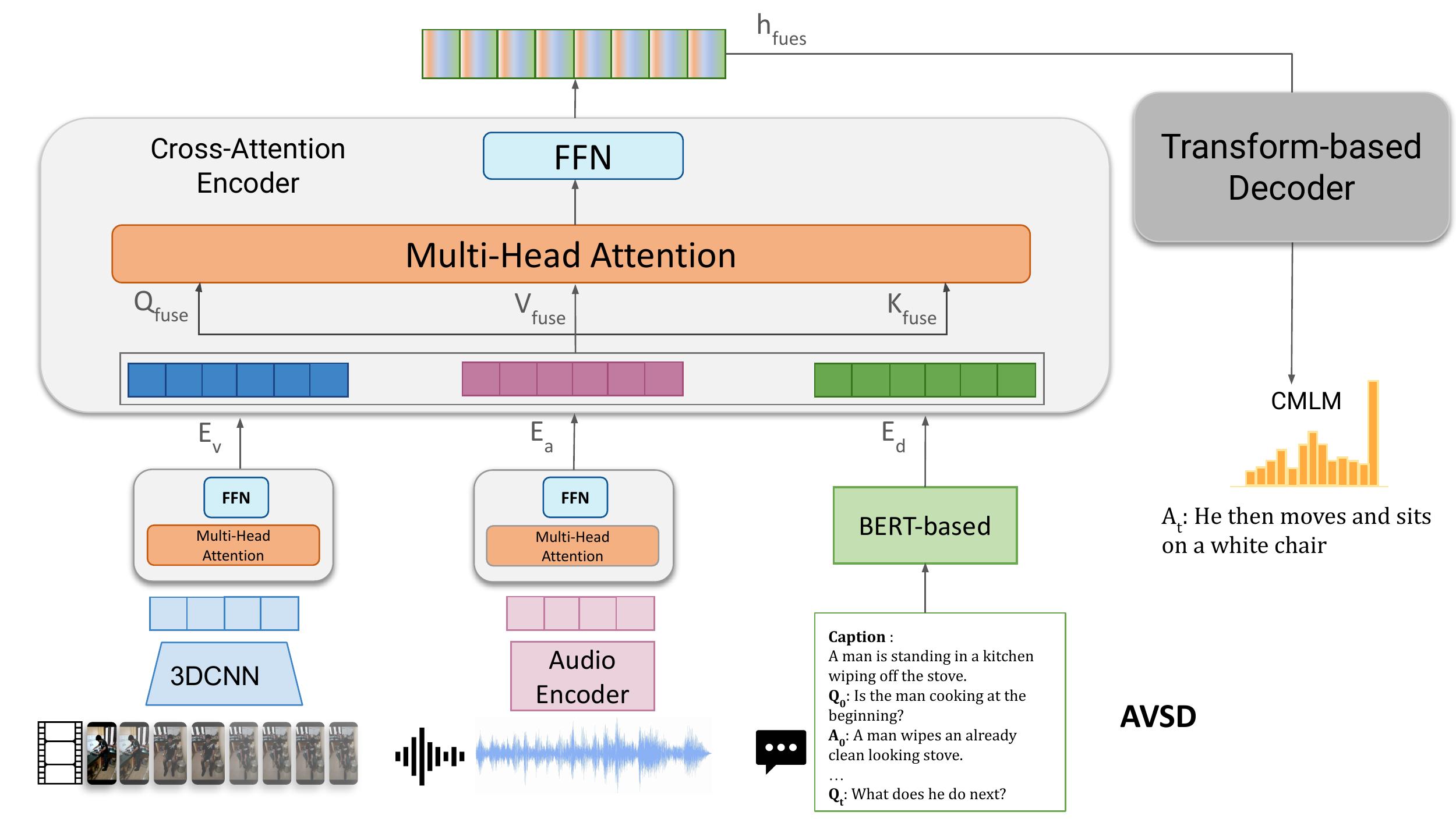}
   \caption{The proposed model consists of \textbf{Visual encoder} that receives sequences of frames and generates an embedding $E_v$, \textbf{Text Encoder} receives text tokens and generates an embedding $E_d$, \textbf{Audio Encoder} generates an embedding $E_a$, \textbf{Multi-Modal Encoder} fuses these embeddings and jointly train the encoders end-to-end.}
\label{fig:framework}
\end{figure*}

\subsection{Model Architecture}


A general overview of the proposed model is presented in Figure~\ref{fig:framework}. It consists of several multi-stream modal specific encoders to extract the initial features, followed by a self-attention network that applies self-attention encoders to generate the contextual representations followed by a transformer-based encoder that generates the final multimodal embedding via cross-attention mechanism.  This is then passed to an auto-regressive decoder to generate an open-ended response.

\vspace{-0.1in}
\subsubsection{Text Encoder}


All the text inputs: $DH$, $C$, $Q$ and $Ans$ are concatenated to form a single long string. Following Devlin~\etal\cite{devlin2018bert} all the words are tokenized using the Word Piece tokenizer~\cite{wu2016google} to obtain a token sequence $t = \left \{ {t_{i}|i \in [1,n]} \right \}$; where  $t_{i}$ is the $i$-th token, and $n$ is the length of the language token sequence. \textsl{[CLS]} token is added at the beginning of the input sequence, and \textsl{[SEP]} is used to separate each sentence (the sentence is either a question or an answer). The processed tokens are then fed to a BERT-based uncased model~\cite{devlin2018bert} to generate a text embedding  $E_{d} \in \mathbb{R}^{n \times d} $, where $d$ is the hidden size of the final self-attention layer of BERT. 
\vspace{-0.5em}
\begin{equation}
  E_{d} = \textsl{BERT}(t)
\end{equation}

\vspace{-0.25in}
\subsubsection{Visual Encoder} 
Initially, a sequence of frames $V_{n} = \left \{ v_{j}  |  j \in  [ 1, m]\right \}$ was subsampled at $16$ fps and cropped to $224$ x $224$. $V_{n}$ is fed into a 3D-CNN network to extract the temporal features. We used \textsl{I3D} network~\cite{I3d} pretrained on ImageNet for the encoder. We extracted global average pooled features from different inception blocks such as $\textsl{Mixed}_4$ and $\textsl{Mixed}_5$ with dimensions $m$ x $d$. Finally, a visual transformer-based encoder applies self-attention over these features $f_{v}$ and generates visual embeddings $E_{v}$. The visual encoder consists of $N=6$ layers of Multi-Head Attention (\textsl{MHA}) and Feed-Forward Network (\textsl{FFN}).

\vspace{-0.1in}
 \begin{equation}
      f_{v} =  \textsl{I3D}(V_{n}), 
 \end{equation}
\vspace{-0.2in}
 \begin{equation}
    E_{v} = \textsl{FFN}(f_{v})+ \textsl{MHA}(f_{v}).
 \end{equation}
     
\vspace{-0.2in}
\subsubsection{Audio Encoder}
To process the audio input, $m$-dimensional features were first extracted using a \textsl{VGGish}~\cite{hershey2017cnn} network.
Similar to the visual features, these were then fed into a transformer-based encoder to extract a contextual representation.
Unlike the visual features, the audio CNN network was not fine-tuned.
\begin{equation}
      f_{a} =  \textsl{VGGish}(A_{m}),
\end{equation}
\vspace{-0.2in} 
\begin{equation}
        E_{a} = \textsl{FFN}(f_{a})+ \textsl{MHA}(f_{a}).
\end{equation}
     
\vspace{-0.2in}
\subsubsection{Cross-Attention Encoder}
Finally, to generate the multimodal representations, we adapted a cross-attention encoder proposed in~\cite{univl} and extended it to learn the interdependencies between the three different modalities. Given the visual $E_v$, audio $E_a$ and dialogue embeddings $E_d$, the encoder fuses them into one sequence and applies cross-attention mechanism over them~\cite{univl}. The cross-encoder consists of $N=6$ \textsl{MHA} layers followed by a Feed Forward Network. 
Finally, a transformer-based auto-regressive decoder is trained to generate responses given the multi-modal representation $h_\textrm{embd}$.

\vspace{-0.1in}
\begin{equation}
      H_{\textrm{fuse}} =  \textsl([ E_{v};E_{a}; E_{d}]),
\end{equation}
\vspace{-0.2in}
\begin{equation}
        h_\textrm{embd} = \textsl{FFN}( H_\textrm{fuse})+ \textsl{MHA}(H_\textrm{fuse}).
\end{equation}

\subsubsection{Training and inference}

The model is trained by optimizing for two objectives losses introduced in~\cite{univl}, namely the Conditioned Masked Language Modeling \textsl{CMLM} and the Decoder Reconstructive Loss. 

For \textbf{\textsl{CMLM}}, \%15 of the input text tokens were masked with \textsl{MASK} special token and the model was trained to predict the masked tokens conditioned on $h_{embd}$.

For the \textbf{Decoder Reconstructive Loss}, at each iteration, the decoder receives the encoded embeddings and generates one answer token ${\hat{y}_{i+1}}$ that is conditioned on the multi-modal fused output and previous generated word ${\hat{y}_{i}}$. 
At inference time, ${\hat{y}_{i+1}}$ with the highest score was chose. 
\begin{equation}
    \hat{y}_{i+1} = argmax P(y_{i+1} = y| \hat{y}_{i},h_{embd}), 
\end{equation}
\vspace{-10pt}

\vspace{-10pt}
\section{Experiments}
\label{sec:experiments}

\subsection{Dataset and evaluation metrics} 
\label{sub3:dataset}

We evaluated our framework on the Audio-Visual Scene-Aware Dialog (AVSD) dataset~\cite{alamri2019audio}. 
It comprises of dialogs grounded in human-based action videos and videos from the Charades~\cite{sigurdsson2016hollywood} dataset. 
Each dialog consists of a video caption and 10 rounds of questions and answers about the events in the video. In total, there are 7,659, 1,787, and 1,710 dialogs in the train, val and test sets respectively.

The results on the DSTC-test set are reported using the common natural language generation 
evaluation metrics including:\textbf{BLEU}~\cite{papineni2002bleu},  \textbf{METEOR}~\cite{banerjee2005meteor}, \textbf{ROGUE-L}~\cite{lin2004rouge}, 
and \textbf{CIDEr}
~\cite{vedantam2015cider}. The test set has only one correct answer for each question.

\subsection{Preprocessing:}

\noindent\textbf{Dialog History:} For the dialog input, we use up to $3$ turns of dialog history with a maximum length of $100$ words, which was generally sufficient for 3 rounds of dialog history plus the question.\newline
\noindent\textbf{Video-Audio features:} We extracted $1024$-d feature from \textbf{$\textsl{Mixed}_{5c}$} and \textbf{$\textsl{Mixed}_{4c}$} layers. 
For comparison with the AVSD baselines~\cite{alamri2019audio}, we pre-extracted the visual features using I$3$D~\cite{I3d} trained on the ImageNet dataset and used the $1024$-d output from the \textbf{$\textsl{Mixed}_{5c}$} layer for the baseline. For the Audio modality, we use pretrained $1024$-d features from $\textsl{VGGish}$~\cite{hershey2017cnn}. This encoder is not fine-tuned on the AVSD dataset.
\vspace{-5pt}
\subsection{Training}
We use the Adam optimizer \cite{kingma2014adam} with a learning rate of $5e^{-5}$ and batch size of $64$.
Training was done using 8 RTX-6000 GPUs.
Early stopping and checkpoint that achieved the best performance on the validation set was selected.

\section{Results and Analysis}
\label{sub3:results}

In this section, we first perform a detailed analysis on the generative task.
For a more well-rounded understanding of the contribution of visual features, several ablations on the retrieval task were performed, -- where the answer is retrieved from a pool of candidate options. 
Finally, we demonstrate the performance of the proposed method via qualitative evaluation.

\begin{table}[h]
\centering
 \caption{Model performance on the AVSD test for the generative task. * includes Audio, $\dagger$ includes summary.}
\renewcommand*{\arraystretch}{1.0}
\resizebox{\columnwidth}{!}{
    \begin{tabular}{l c c c c c c c c }
    \toprule
      \textbf{Method} & \textbf{BLEU2}$\uparrow$  & \textbf{BLEU3}$\uparrow$  & \textbf{BLEU4}$\uparrow$  & \textbf{METEOR}$\uparrow$  & \textbf{ROGUE-L}$\uparrow$ & \textbf{CIDEr}$\uparrow$\\
    \midrule
      DGR* (2021) & -- & -- & 0.357 & 0.267 & 0.553 & 1.004 \\ 
      JST*$\dagger$(20219) & {--} & {--} & 0.406 & 0.262 & 0.554 & 1.079 \\
      VideoGPT2*$\dagger$ (2020) & 0.570 & 0.476 & 0.402 & 0.254 & 0.544 & 1.052 \\
      MTN $\dagger$ (2019) & 0.242 & 0.174 & 0.135 & 0.165 & 0.365 &  \textbf{1.366}\\
      \midrule
      JMAN (2020)\cite{chu2020multi} & 0.521 & 0.413 & 0.334 & 0.239 & 0.533 & 0.941 \\
      Le H. et. al.(2021)\cite{le2021c}  & 0.577 & 0.476 & 0.398 & 0.262 & 0.549 & 1.040 \\
      TimeSformer *$\dagger$ (2022)~\cite{yamazaki2022audio} & 0.572 & 0.477 &  0.403 & 0.255 & 0.547 & 1.049 \\
      \bottomrule
      Ours + Audio modality*  & 0.587 & 0.483 & 0.401 & \textbf{0.271} & 0.565 & 1.155 \\
      Ours  & \textbf{0.592} & \textbf{0.493} & \textbf{0.415} & 0.269 & \textbf{0.569} & 1.159 \\
      \bottomrule
      \end{tabular}}
    \label{tab:modelPerf}
\end{table}

\vspace{-0.1in}
\subsection{Results on the Generative task}
We compare our results with ~\cite{le2019multimodal,videoGPT2,shah2021audio}. VideoGPT~\cite{videoGPT2} uses GPT2\cite{GPT2}, a pretrained generative encoder that is known to outperform BERT\cite{bert18} model that we adapted. JST and MTN are also self-attention based models, however they do not finetune the visual backend network to AVSD dataset and feed pre-extracted visual features. 
Table \ref{tab:modelPerf} Shows that our model outperforms these models across the different evaluation metrics, achieving a gain in 
BLUE2 (0.592 \textendash\textgreater 0.570),
BLUE3 (0.493\textendash\textgreater 0.476), 
BLUE4 (0.415\textendash\textgreater 0.406),
METEOR (0.269 \textendash\textgreater 0.267) 
and ROUGEL (0.569 \textendash\textgreater 0.356). 
For CIDEr, although our method underperforms to the MTN method, the latter uses a much larger model and more textual input (Summary, in addition to caption, and dialog history). 
These results indicate that the joint training improves the model`s utilization of the visual features, and with only a slight increase in memory and time cost, performs better or comparable results to models with deeper networks.
By using standard architectures, we highlight the gains due to the textual-visual association rather than stronger language encoder that is not visually-aware. We would like to reiterate that the novelty of the proposed work lies in the approach taken in learning the joint features, and the performance improvement achieved speaks to that.
\subsubsection*{Role of Audio Modality:}
In Table~\ref{tab:modelPerf}, we show methods that use Audio with $^*$. When compared to our method that uses Audio and Text inputs, using Audio does not show a significant improvement.
This is because, for AVSD dataset specifically, the audio has sounds without any dialog. However, for completeness, and generalisation to other datasets, we have included the results in the table.

\vspace{-0.1in}
\subsection{Results on the Retrieval task}
\label{sub:retrieval}
To establish that visual information aids effective performance in the AVSD task, we further evaluate our proposed approach on the retrieval setting. In the retrieval setting, the model is given the same inputs as the generative task but tasked with retrieving the correct answer from a pool of candidate answers by outputting a ranking. This settings allows for direct evaluation of the encoded modalities without the decoder performance.

\begin{figure*}[t]
    \centering
    \includegraphics[width=0.55\textwidth]{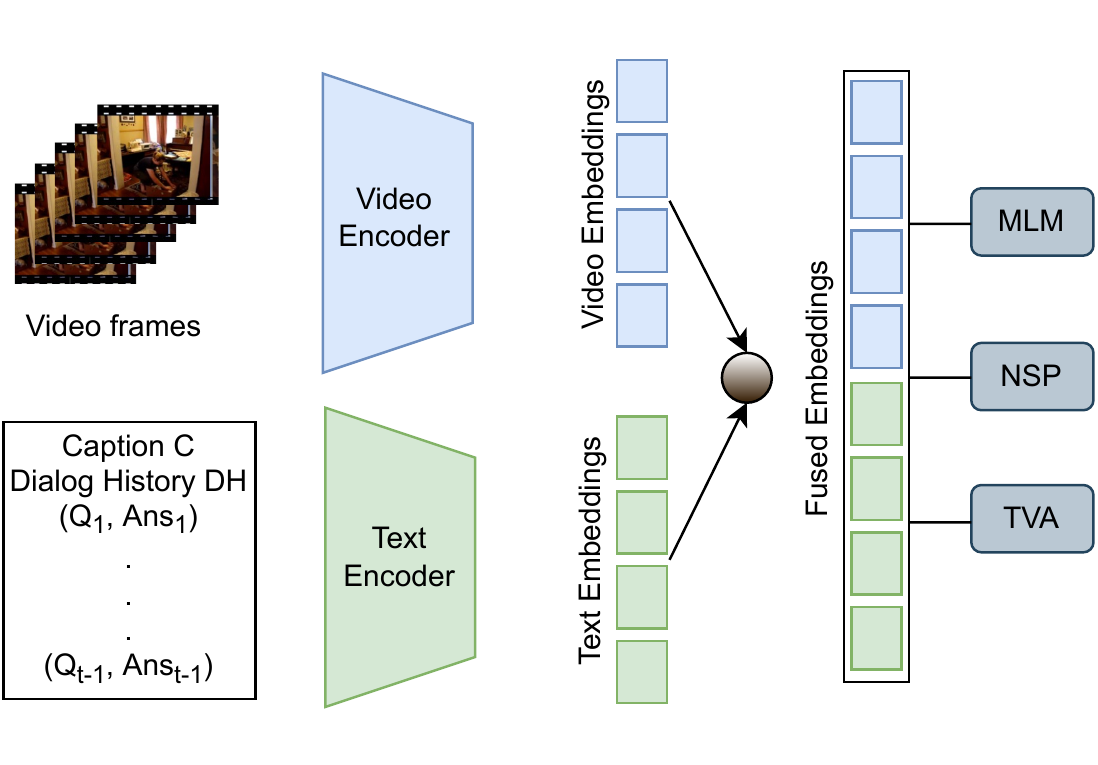}
   \caption{Retrieval task: The classification model consists of Visual encoder that receives sequences of frames and generates an embedding $E_v$,Text Encoder receives text tokens and generates an embedding $E_d$,Dialog Encoder fuses these embeddings and jointly train the encoders end-to-end.} 
\label{fig:framework_cvpr}
\end{figure*}

For this purpose, we design a much simpler framework as shown in \ref{fig:framework_cvpr}, where the video embeddings from I$3$D and text embeddings from the BERT model are concatenated and optimized for the following objectives: Masked Language Model loss~($L_\text{mlm}$), Next Sentence Prediction loss~($L_\text{nsp}$), and text-video alignment loss~($L_\text{vta}$). 
The retrieval model is also trained jointly, thus learning from the visual-text association.
The training objectives are detailed in the Appendix~\ref{sec:appendix}. 

\begin{table*}[th!]
\begin{center}
\caption{Model performance on the AVSD dataset. $\text{XXX}_\text{ft}$ refers to finetuned models,  $\text{XXX}_\text{no-ft}$ to non-finetuned models. $\uparrow$ implies higher the score the better, $\downarrow$ implies, lower the score the better.}
\vspace{1pt}
\setlength{\tabcolsep}{6pt} 
\renewcommand{\arraystretch}{1.2}
\resizebox{0.9\columnwidth}{!}{
\begin{tabular}{l l l c c c c c }
 \toprule
    \textbf{Input} & \textbf{Text Encoder} & \textbf{Vid Encoder} & $\uparrow$ \textbf{MRR} & $\uparrow$ \textbf{R@1} & $\uparrow$ \textbf{R@5} & $\uparrow$ \textbf{R@10} & $\downarrow$ \textbf{MR} \\
 \midrule
    DH & {LSTM} & {-} & 50.40 & 32.76 & 73.27 & 88.60 & 4.72 \\
    & {BERT} & {-} & 69.71 & 56.93 & 86.18 & 92.93 & 5.07  \\
\midrule  
    DH + V & {LSTM} & $\text{I3D}_\text{no-ft}$ & 53.41  & 36.22 & 75.86 & 89.79 & 4.41  \\
    & {LSTM} & $\text{S3D}_\text{no-ft}$ & 53.57 & 36.49 & 75.64 & 89.82 & 4.45  \\
    & {LSTM} & $\text{I3D}_\text{ft}$ & 54.28  & 37.12 & 76.62 & 90.23 & 4.33  \\
\midrule
    \textbf{DH + V (Ours)}  
    & $\text{BERT}_\text{ft}$ & $\text{S3D}_\text{no-ft}$ &  71.32 & 59.51 & 86.92 & \textbf{95.22} & 4.89 \\
    & $\text{BERT}_\text{ft}$ & $\text{S3D}_\text{ft}$ & \textbf{77.28} & \textbf{67.28} & \textbf{90.39} & 94.87 & \textbf{4.18} \\
  \bottomrule
\end{tabular}}
\end{center}
\label{tab:model_perf_retrieval}
\end{table*}

\vspace{1em}
\textbf{Evaluation Metrics}
We report the retrieval metrics: R@1, R@5, R@5, as well as the Mean Rank (MR), and Mean Reciprocal Rank (MRR). Ideally the ground truth answer is ranked first. 


\vspace{1em}
\textbf{Performance of the language encoders:}
Table \ref{tab:model_perf_retrieval} summarizes the results of the evaluation phase. 
For the text encoders, we note that BERT significantly outperforms the LSTM encoder, achieving $69.71$ MRR, which is a $19\%$ absolute improvement over the LSTM-based encoder. 
This development was anticipated as Transformer-based encoders, such as BERT, benefit from pre-training with a large text corpora on several proxy tasks. 
This can generate a rich contextualized representation that assists the model to understand linguistic input.
However, we would like to note that even when a simple language encoder such as LSTM is used, jointly training the visual encoder results in a significant performance improvement.

\vspace{1em}
\textbf{Effect of different visual encoders:} We measure the effect of utilizing different $3$D-CNN networks to extract the visual representations, and observe that they perform comparably, achieving $53.41$ MRR for the I$3$D features, and $53.57$ for S$3$D (\ref{tab:model_perf_retrieval}). 
This indicates that alterations to the visual encoder do not improve the model performance, as the actual joint training for the network was the chief factor for the performance.

\vspace{1em}
\textbf{Joint training is effective:}
Combining the visual features with the fine-tuned language BERT encoder - $\text{BERT}_\text{ft} + \text{S3D}_\text{no-ft}$, outperforms the language only model with an increase of $1.6\%$ on MRR. 
This modest improvement when adding visual features has been the main trend in video-based dialog systems \cite{alamri2019audio,hori2019end,videobert,dstc8}. 
Finally, the model that is jointly fine-tuned on both modalities achieved the best performance across all metrics, with a \textbf{$6\%$} increase in MRR.

\subsection{Effect of the number of fine-tuned blocks.}
\label{blocks}
In Table~\ref{tab:inc_blocks} we detail the impact of fine-tuning several inception blocks on our visual encoders. The S3D network comprises three convolution layers, followed by five inception blocks. The depth and network architecture resulted in additional trainable parameters. The aim was to learn the effect of conducting fine-tuning across more layers, i.e. inception blocks, to ascertain whether this would allow the model to generate better visual features for the task. Table~\ref{tab:inc_blocks} shows the model’s performance when fine-tuning different inception models.

\begin{table}[h!]
\begin{center}
\caption{Performance by finetuning different inception blocks from the visual encoder} 
\setlength{\tabcolsep}{6pt} 
\renewcommand{\arraystretch}{1.2}
\resizebox{1\columnwidth}{!}{
\begin{tabular}{l c c c c c c c c c c c c}
\toprule
 & \multicolumn{5}{c}{Retrieval Mode} &  & \multicolumn{6}{c}{Generative Mode} \\ \cline{2-6} \cline{8-13}\\
{Trained Inception Blocks} & $\uparrow${MRR} & $\uparrow$ {R@1} & $\uparrow${R@5} & $\uparrow${R@10} & $\downarrow${MR} & & $\uparrow${BLEU2} & $\uparrow${BLEU3} & $\uparrow${BLEU4}  & $\uparrow${METEOR} & $\uparrow${ROGUE-L} & $\uparrow${CIDEr} \\
\cmidrule{2-6} \cmidrule{8-13}
$\text{S3D}_\text{no-ft}$ & 53.41  & 36.22 & 75.86 & 89.79 & 4.41 & & 0.58 & 0.488 & 0.407 & 0.268 & 0.561 & 1.115 \\
$\text{S3D}_{\text{Mixed}\-5}$ & \textbf{77.21} & \textbf{67.20} & 90.22 & \textbf{95.06} & \textbf{4.15} & & \textbf{0.592} & 0.492 &	0.413 & 0.267 &	0.563 &	1.134 \\
$\text{S3D}_{\text{Mixed}\-4, \text{Mixed}\-5}$ & 76.88 & 66.72 & \textbf{90.39} & 94.77 & 4.48 & & \textbf{0.592}& \textbf{0.493} & \textbf{0.415} & \textbf{0.269} & \textbf{0.569} & \textbf{1.159} \\
\bottomrule
\end{tabular}}
\end{center}
\label{tab:inc_blocks}
\end{table}

\subsection{More frames are more informative:} 
\label{frames}
We also conducted an experiment in which we varied the size of the sampled frames. The question we are seeking to answer was: \textit{how many video frames are sufficient to answer the input question?} We trained the model using sampling frame sizes: $6$, $16$, $32$ and $40$. As presented in Table~\ref{tab:moreframes}, we can see that the model performance improved significantly when trained on larger sampling rates, concluding that the model benefits greatly from additional visual features when trained jointly on downstream tasks. We see a small drop in performance at frame rate of $40$ frames. We believe this is because the pre-trained model was trained with $30$ frames per video sequence, and an increasing the number of frames results in redundant data.



\subsection{Dialog history is helpful:} 
\label{dialog_history}
We evaluated the effect of the length of the dialog information. We tested the model performance in the first round -$\text{Round}_1$, where there were no prior dialog utters, and in $\text{Round}_{3}$, where there were 2 previous dialog utters, increasing in $\text{Round}_5$ and $\text{Round}_{10}$. The results for each round are displayed individually in the AVSD test set in~\ref{tab:dh_length}. As we can observe, the model performance improved from the first round, with $59.57$ MRR, to the third round, which obtains $81.66$ MRR. This was because the third round included information from previous rounds. As the dialog tends to become more generic and uninformative after the initial Q and As as seen  Figure~\ref{fig:example_QA-2}, we see a performance drop after the third round. 

\begin{table}[h]
\begin{minipage}[c]{0.545\textwidth}
\caption{Evaluation results for on the test set of AVSD Performance by total number of sampled video frames} 
\centering
\setlength{\tabcolsep}{6pt} 
\renewcommand{\arraystretch}{1}
\resizebox{1\columnwidth}{!}{
\begin{tabular}{c c c c c c}
\toprule
\begin{tabular}[c]{@{}c@{}}Number of \\ Sampled Frames\end{tabular}  & $\uparrow${MRR} & $\uparrow$ {R@1} & $\uparrow$ {R@5} & $\uparrow$ {R@10} & $\downarrow$ {MR} \\
\midrule
6  & 46.38 & 31.15 & 64.70 & 78.87 & 8.46 \\
16 & 74.90 & 64.11 & 89.19 & 94.47 & 4.63 \\
32 & \textbf{77.28} & \textbf{67.92} & \textbf{90.22} & \textbf{95.06} & \textbf{4.15} \\ 
40 & 77.21 & 66.20 & 89.62 & 94.82 & 4.46 \\ 
\bottomrule
\end{tabular} 
\label{tab:moreframes}
}
\end{minipage}
\hspace{0.1in}
\begin{minipage}[c]{0.425\textwidth}
\caption{Evaluation of length of the dialog history on the Performance.} 
\centering
\setlength{\tabcolsep}{6pt} 
\renewcommand{\arraystretch}{1}
\resizebox{1\columnwidth}{!}{
\begin{tabular}{c c c c c c}
\toprule
\begin{tabular}[c]{@{}c@{}}Dialog \\ Round\end{tabular} & {MRR} & {R@1} & {R@5} & {R@10} & \begin{tabular}[c]{@{}c@{}}Mean \\ rank\end{tabular} \\
\midrule
1 &  59.57  & 46.39 & 75.70 & 87.71 & 4.49  \\
3 & \textbf{81.66} & \textbf{71.65} & \textbf{94.82} & \textbf{98.81} & \textbf{1.82} \\
5  & 77.21 & 67.20 & 89.62 & 94.82 & 4.46\\ 
10 & 62.52 & 52.96 & 74.46 & 78.77 & 18.34\\ 
\bottomrule
\end{tabular} 
\label{tab:dh_length}
}
\end{minipage}
\centering
\end{table}

\begin{figure*}
\centering
         \includegraphics[width=0.6\columnwidth,height=0.25\columnwidth]{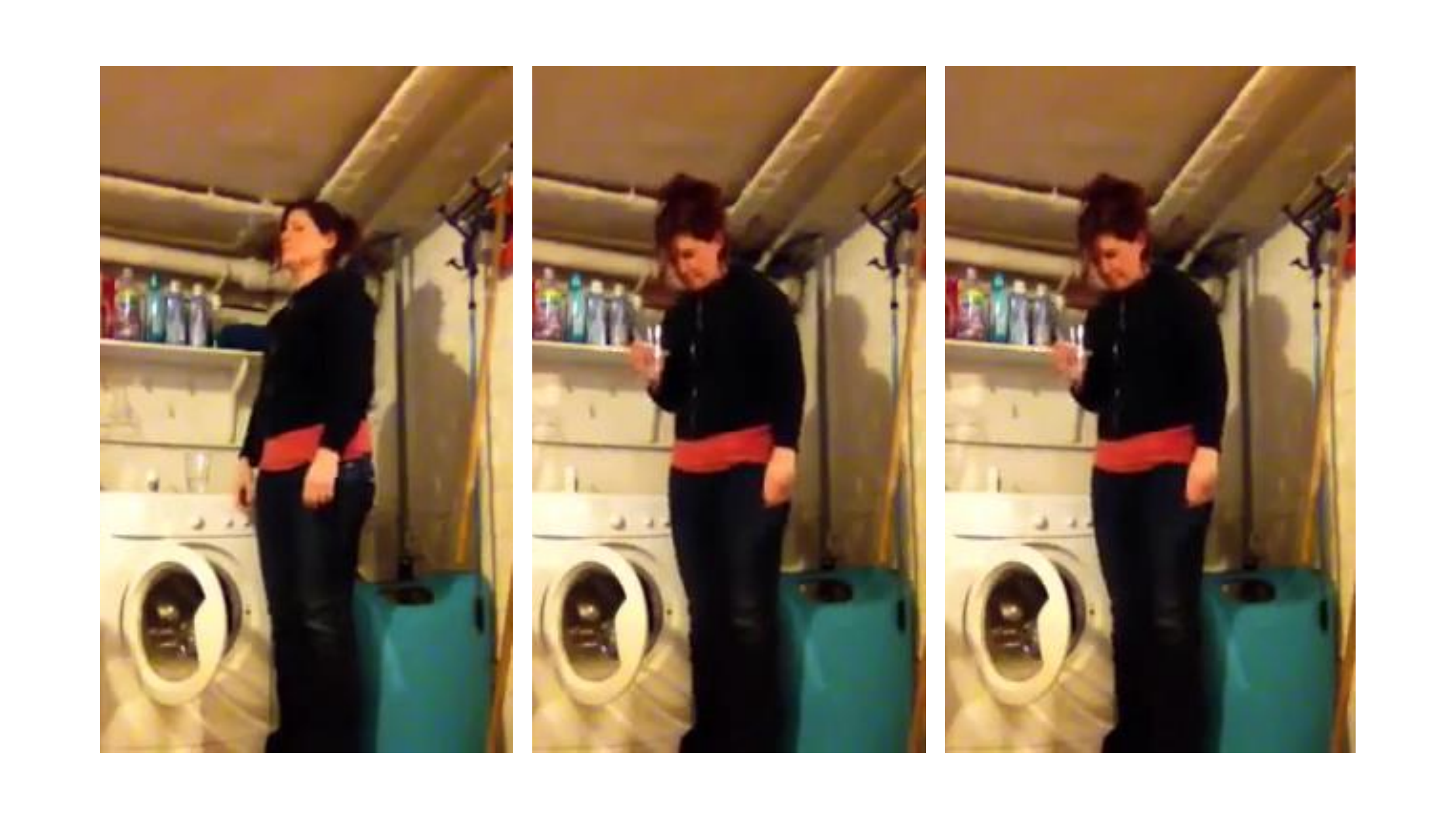} 
\centering
\resizebox{0.9\columnwidth}
{!}{
\setlength{\tabcolsep}{6pt} 
\renewcommand{\arraystretch}{1.2}
\begin{tabular}{p{0.05\linewidth} p{0.5\linewidth} p{0.50\linewidth}}
\hline
\textbf{Id} & \textbf{Question}  & \textbf{GT\_Answer} \\ \toprule
\textbf{0} & what is the person looking at in the beginning?  & she is looking at the glass of water in her hands \\ \hline
\textbf{1} & Is she in the laundry room? & Yes, it looks to be in the basement of her home \\ \hline
\textbf{2} & Is she doing laundry at all & No, she sneezes and takes some medicine  \\ \hline
\textbf{3} & Where does she get the medicine from? & A bottle on the washing machine \\ \hline
\textbf{4} & Does she put the bottle back on washing machine? & Yes, she does, then she drinks the water \\ \hline
\textbf{5} & Does she set the glass down? & She sets the glass down after sneezing to get the medicine  \\ \hline
\textbf{6} & Does she ever move around? & No she stays in the same place   \\ \hline
\textbf{7} & Does the video end with her drinking water & Yes she is drinking the water at the end. \\ \hline
\textbf{8} & Does she say anything at all & No she does not speak. \\ \hline
\textbf{9} & Are there any noises in video  & Only her sneeze can be heard chips \\  
\bottomrule
\end{tabular}}


\caption{Questions and answers in a typical dialog setting. We see that the first few questions are closely related to the video, but the later ones are very generic which makes it hard for the BERT model to train on.\label{fig:example_QA-2}}
\end{figure*}

\section{Conclusion}
\label{sec:conclusion}

In this paper, we proposed a new framework for a video-based dialog task. In our framework we optimized the learning from the visual input by jointly training the visual encoder end-to-end with different modalities like text and audio. 
Different generative and retrieval tasks showed that our training scheme generates a more rich multimodal representation and helps reduce the bias towards the textual information. 
We emphasize that \textbf{joint learning of visual and textual information is vital} for the video dailog task.
Though there is an additional time cost and memory cost, our results show a significant improvement across all tasks and metrics, thus re-enforcing our belief that the finetuning the video encoders is crucial for the tasks. 
Future work will aim to extend our goal to pretrain our model using self-supervision tasks in raw unlabeled data then use the generated representations for more complex tasks such as video captioning, and video retrieval.

\clearpage
{
\small
\bibliographystyle{ieee_fullname}
\bibliography{egbib}
}
\clearpage
\appendix
\section*{Appendix}
\label{sec:appendix}
\section{Retrieval task: Implementation details}

\subsection{Text Encoder}

To generate the text representation we utilize the BERT-based uncased model~\cite{devlin2018bert}. We feed the processed tokens to the text encoder to generate a text embedding $T_{d} \in \mathbb{R}^{n \times d} $, where $d$ is the hidden size of the final self-attention layer of BERT. 

\begin{figure*}[h!]
    \centering
    \includegraphics[width=0.9\textwidth, height= 6 cm]{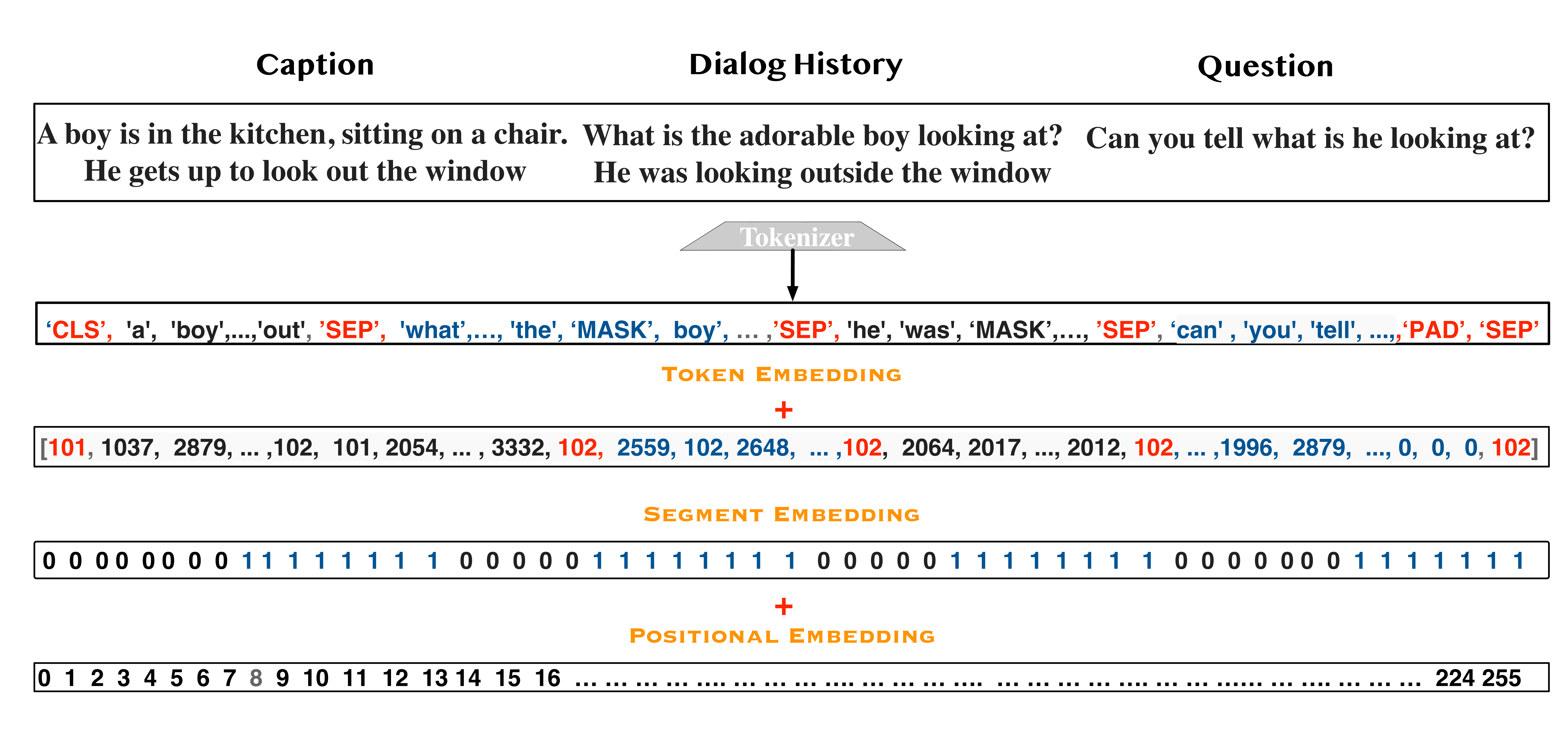}
   \caption{The final input for the text encoder is the sum of the token embedding, segment embedding and the position encoding} 
\label{fig:embeddings}
\end{figure*}

We concatenate \texttt{DH}, \texttt{C},\texttt{C} and \texttt{Q} to form a single long string. Next, we follow Devlin J~\etal\cite{devlin2018bert} and tokenize all the words using the Word Piece tokenizer~\cite{wu2016google} to obtain a token sequence $t = \left \{ {t_{i}|i \in [1,n]} \right \}$; where  $t_{i}$ is the $i$-th token, and $n$ is the length of the language token sequence. $<CLS>$ token is added at the beginning of the input sequence, and $<SEP>$ is used to separate each sentence (the sentence is either a question, or an answer). In addition to the embedding of these words, we add positional embedding, and segment embedding. For the segment embedding we followed~\cite{visdialbert} and added additional segment embeddings for the questions and answers, see Figure~\ref{fig:embeddings}.

The hidden size of the model is 768 and the batch size is 16. For the dialog input, we used up to $3$ turns of dialog history with a maximum length of $200$ words.

\subsection{Visual Encoder} 
For the visual encoder, we use S3D~\cite{xie2017rethinking}, which is a 3DCNN built on the Inception network~\cite{szegedy2015going} as a backbone with separable temporal convolutions.

First, we subsample a sequence of frames  $v = \left \{ v_{j}  |  j \in  [ 1, m]\right \}$ at the rate of $m$ fps and $224$ x $244$ frame size. Next, we feed the frames sequence to the pretrained S$3$D and extract global average pooled features from different inception blocks such as $Mixed\_4$ and $Mixed\_5$. The extracted features, $V_{e}$ have the dimension $m$ x $1024$. We present ablations over different frame rates in Section~\ref{frames}. 

\subsection{Training Objectives} 
\label{End-to-End training}
We train our model jointly end-to-end by optimizing for the following objectives: Masked Language Model loss ($L_\text{mlm}$), Next Sentence Prediction loss ($L_\text{nsp}$), and text-video alignment loss ($L_\text{vta}$). MLM and NSP tasks are widely used to perform transfer learning and fine tuning of pre-trained Transformer-based models in new downstream tasks~\cite{devlin2018bert, attention}. In our work we adopted those tasks as follows: 

\begin{itemize}
\item For the Masked Language Modelling task, we masked 10\% of the final input tokens and replaced those tokens with a \texttt{<MASK>} token; and trained the model to predict the masked tokens from the surrounding ones.  

\item In the Next Sentence Prediction task, given two sentences \texttt{A} and \texttt{B} a label \textbf{O}, we trained the model to give an output of \textbf{1} if \texttt{A} and \texttt{B} are related and should appear together, and output \texttt{0} otherwise. \texttt{A} is the concatenated input tokens: \texttt{$C+DH_{t} +Q_{t}$}, and \texttt{B} represents the ground-truth answer for a positive example. For a negative example, we randomly select the sample from the list of candidate answers. 
\item Text-video alignment task: After the visual embedding $V_e$ is extracted using the aforementioned visual encoder, we apply a fully-connected layer to transfer $V_e$ to the same dimensional space at $T_e$. Then we concatenate these representations to get the fused token embedding $\text{fused}_e$. Then we apply inner product between $\text{fused}_e$ and a candidate answer embedding $a_e$, and train the model with negative log-likelihood and \textit{k}-negative samples with the weighted total losses: $L_\text{mlm}$, $L_\text{nsp}$ and $L_\text{vta}$. 
 
\end{itemize}

\subsection{Qualitative results:}
\begin{figure*}[h]
\centering
\includegraphics[width=0.5\columnwidth,height=0.2\columnwidth]{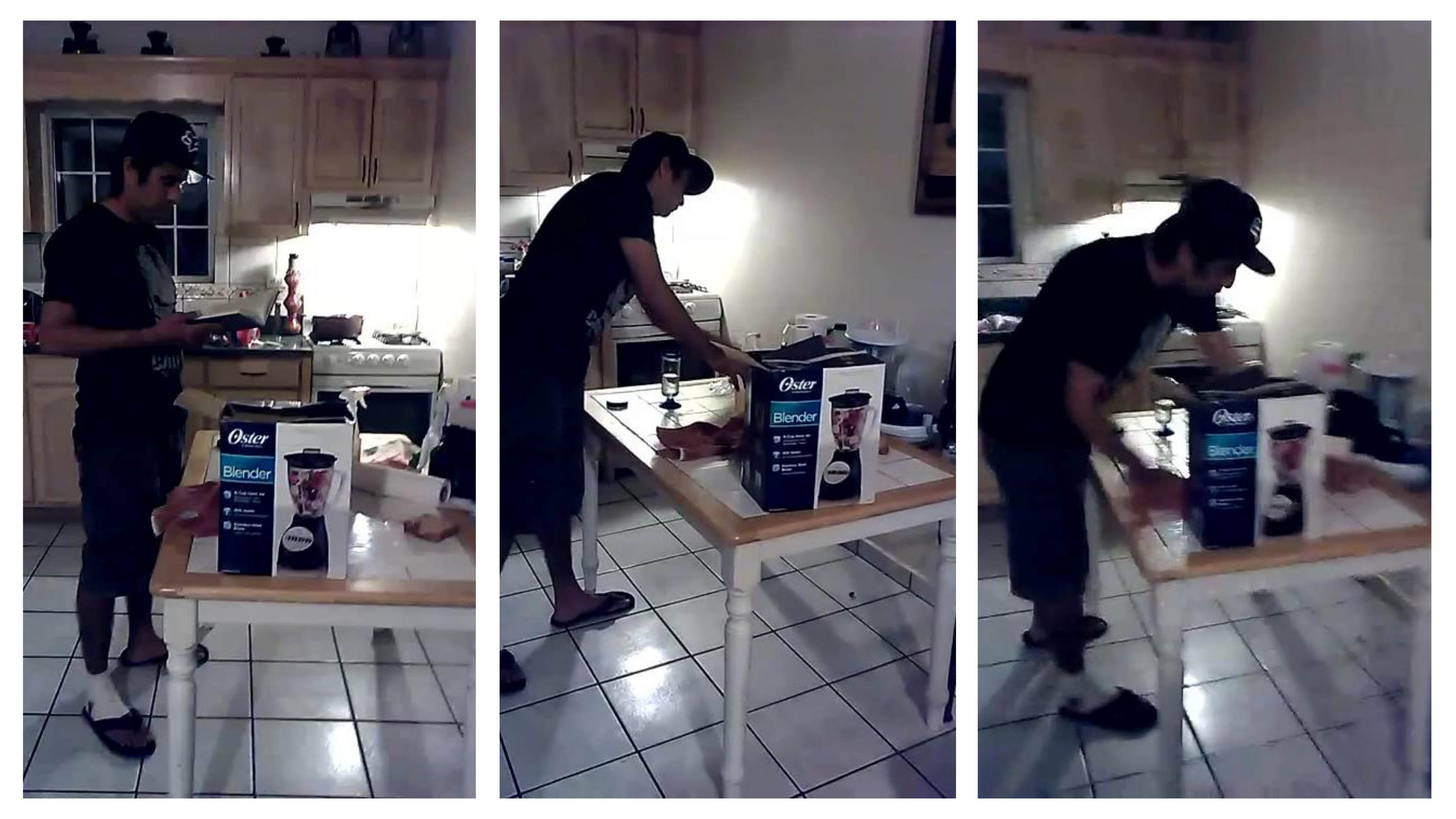}  
\small
\resizebox{0.95\columnwidth}
{!}{
\setlength{\tabcolsep}{6pt} 
\renewcommand{\arraystretch}{1}
\begin{tabular}{p{0.25\linewidth}  p{0.35\linewidth} p{0.35\linewidth}
p{0.25\linewidth}}
\toprule
 \textbf{Question} &
  \textbf{GT\_Answer}  &
  \textbf{Top\_Predicted\_Answer} &
  \textbf{Top\_Predicted\_Answer} \\
  & &
  $\text{BERT}_\text{ft} + \text{S3D}_\text{no-ft}$ &
  $\text{BERT}_\text{ft} + \text{S3D}_\text{ft}$ \\
\midrule
  {\color[HTML]{000000}Is there only one man in the video?} &
  {\color[HTML]{000000}Yes, he is the only person in the video} &
  \color[HTML]{009901}Yes, just the one man. &
  \color[HTML]{009901}There is one person in the video. \\ \hline

  {\color[HTML]{000000}In the beginning, is he looking at a book?} &
  {\color[HTML]{000000}Yes, he is flipping though it before tossing it into the box} &
  {\color[HTML]{CB0000} Yes it is, might be checking for something.} &
  {\color[HTML]{009901} Yes, he is flipping though it before tossing it into the box} \\ \hline

  {\color[HTML]{000000}Do you think it is an instruction manual?} &
  {\color[HTML]{000000}It may have been, it didn't look very thick }&
  {\color[HTML]{CB0000} Yes probably. like a tall hair} &
  {\color[HTML]{009901} It may have been, it didn't look very thick} \\ \hline

  {\color[HTML]{000000}Does he toss it in the blender box? }&
  {\color[HTML]{000000}Yes, in the blender box }&
  {\color[HTML]{009901} Yes, both the kitchen and the living room can be seen in the video.} &
  {\color[HTML]{009901} Yes, in the blender box} \\ \hline

  {\color[HTML]{000000}What happens next in the video?} &
  {\color[HTML]{000000}He starts wiping the island top off with a cloth} &
  {\color[HTML]{CB0000} Yes, probably. like a tall hair} &
  {\color[HTML]{CB0000} Yes, probably. like a tall hair} \\ \hline

  {\color[HTML]{000000}Is he using a cleaner or just a cloth?} &
  {\color[HTML]{000000}He is only using the cloth} &
  {\color[HTML]{009901}The is only using the cloth} &
  {\color[HTML]{009901}The is only using the cloth} \\ \hline

  {\color[HTML]{000000}What happens after that? }&
  {\color[HTML]{000000}Nothing. The video ends with him still wiping the island top off} &
  {\color[HTML]{009901}Nothing. The video ends with him still wiping the island top off} &
  {\color[HTML]{009901}Nothing. The video ends with him still wiping the island top off} \\ \hline

 {\color[HTML]{000000}So does he ever leave the room?} &
 {\color[HTML]{000000}No, he stays in the kitchen} &
 {\color[HTML]{CB0000}Yes, at the very end} &
 {\color[HTML]{CB0000}Yes, at the very end} \\\hline

 {\color[HTML]{000000}No one else enters the room?} &
 {\color[HTML]{000000}No, there is no one else in the room} &
 {\color[HTML]{CB0000} There is only one person} &
 {\color[HTML]{CB0000} There is only one person} \\ \hline

 {\color[HTML]{000000}Does he ever look directly at the camera?} &
 {\color[HTML]{000000}No, only at the book and the island top }&
 {\color[HTML]{009901} No, only at the book and the island top} &
 {\color[HTML]{009901} No, only at the book and the island top} \\ \bottomrule
\end{tabular}}

\caption{Example $1$: We show the answers generated by our model (Jointly trained BERT $+$ Video Encoder) and compare it with the separately trained BERT $+$ Video Encoder. The answers in {\color[HTML]{009901}green} are the correct answers while {\color[HTML]{CB0000}red} are incorrect answers generated by the models}
\label{fig:qualitative-1} 
\end{figure*}

\begin{figure*}[t!]
\centering
\includegraphics[width=0.7\columnwidth,height=2cm]{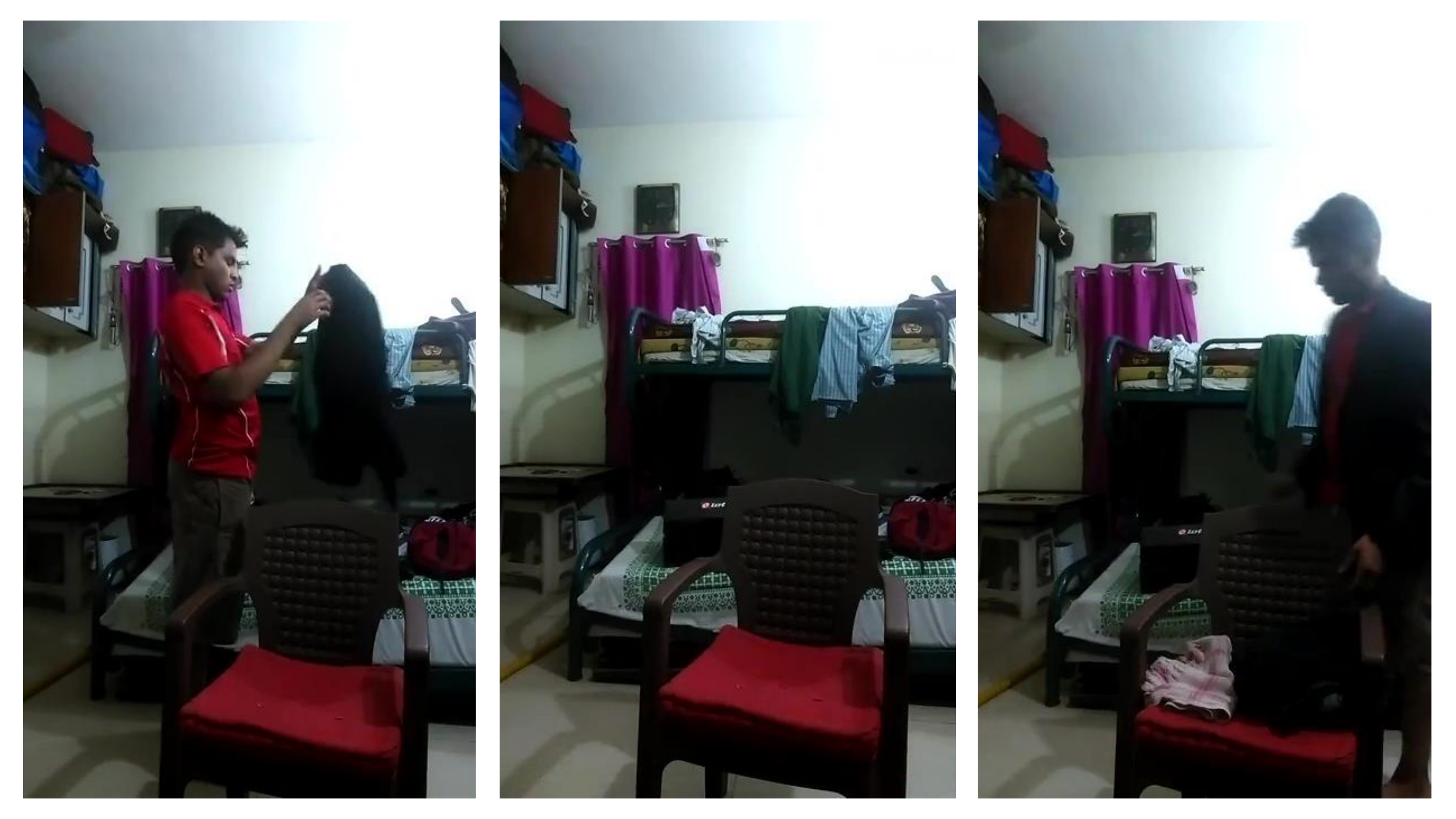}  
\small
\resizebox{0.95\columnwidth}{!}{
\setlength{\tabcolsep}{6pt} 
\renewcommand{\arraystretch}{1.1}
\begin{tabular}{p{0.01\linewidth} p{0.25\linewidth} p{0.25\linewidth} p{0.35\linewidth} p{0.35\linewidth}}
\toprule
 &
  \textbf{Question} &
  \textbf{GT\_Answer}  &
  \textbf{Top\_Predicted\_Answer} &
  \textbf{Top\_Predicted\_Answer} \\
  & & &
  $\text{BERT}_\text{ft} + \text{S3D}_\text{no-ft}$ &
  $\text{BERT}_\text{ft} + \text{S3D}_\text{ft}$ \\
\midrule
0 &
  Where does the video take place? &
  Looks like a bedroom maybe &
  {\color[HTML]{CB0000} It seems to start in a pantry.} &
  {\color[HTML]{009901} Looks like a bedroom maybe} \\ \hline
1 &
  How many people are in the video? &
  There is just one person &
  {\color[HTML]{CB0000}One older gentleman is the person in the video} &
  {\color[HTML]{CB0000}One older gentleman is the person in the video} \\ \hline
2 &
  What gender does the person in the video appear to be? &
  The person in the video is male &
  {\color[HTML]{009901}The person in the video is male} &
  {\color[HTML]{009901}The person in the video is male} \\ \hline
3 &
  How old does he appear to be? &
  Not sure, maybe in his twenties &
  {\color[HTML]{CB0000}He looks young, id guess teens maybe} &
  {\color[HTML]{009901}Not sure, maybe in his twenties} \\ \hline
4 &
  What happens after he enters the room? &
  He grabs a towel and some clothes and then goes off screen &
  {\color[HTML]{009901}He grabs a towel and some clothes and then goes off screen} &
  {\color[HTML]{009901}He grabs a towel and some clothes and then goes off screen} \\ \hline
5 &
  Does he place the towel and clothes anywhere? &
  Not that i can see because he goes off screen with them &
  {\color[HTML]{009901}Not that i can see because he goes off screen with them} &
  {\color[HTML]{009901}Not that i can see because he goes off screen with them} \\ \hline
6 &
  Does he ever come back on screen? &
  Yes he comes back without the stuff and then starts taking his jacket off &
  {\color[HTML]{CB0000}No he stays in that room for the rest of the clip} &
  {\color[HTML]{009901}Yes he comes back without the stuff and then starts taking his jacket off} \\ \hline
7 &
 Where does he place his jacket? &
 He carefully places it on the bottom bunk of the bed &
 {\color[HTML]{CB0000} He is putting them on counters it looks like} &
 {\color[HTML]{009901} He carefully places it on the bottom bunk of the bed} \\ \hline
8 &
 How does the video end? &
 He goes to his belt like he is going to take it off but he is kind of off screen and then it ends &
{\color[HTML]{009901} He goes to his belt like he is going to take it off but he is kind of off screen and then it ends} &
 {\color[HTML]{009901} He goes to his belt like he is going to take it off but he is kind of off screen and then it ends} \\ \hline
9 &
 Did you hear any sounds in the video? &
 No I don't hear anything in the video &
 {\color[HTML]{CB0000} No I heard no speaking} &
 {\color[HTML]{009901} No I don't hear anything in the video} \\ \bottomrule
\end{tabular}}

\caption{We show the answers generated by our model (Jointly trained BERT $+$ Video Encoder) and compare it with the separately trained BERT $+$ Video Encoder. The answers in {\color[HTML]{009901}green} are the correct answers while {\color[HTML]{CB0000}red} are incorrect answers generated by the models.}
\label{fig:qualitative-2}
\end{figure*}

We show qualitative results from our retrieval model in Figure.~\ref{fig:qualitative-1} and Figure.~\ref{fig:qualitative-2}. We see that our model is able to retrieve better answers compared to the video encoder is not finetuned to the task. As presented in these example, our best model ranks the correct answer as the top predicted answer more frequently than a the baseline model, with the same language encoder. Indicating that the model benefits largely from the joint end-to-end training of the visual and language encoders.

\vspace{0.2em}

\end{document}